%% file: itsc.tex
\documentclass[letterpaper, 10 pt, conference]{ieeeconf}  

\IEEEoverridecommandlockouts                              

\usepackage{subfigure}
\usepackage{graphicx}
\usepackage{url}
\usepackage{amsmath}
\usepackage{amssymb}
\usepackage{caption}
\usepackage{algorithmic}
\usepackage{algorithm}
\usepackage{color}
\usepackage{float}
\usepackage{listings}
\usepackage[utf8]{inputenc}
\usepackage[english]{babel}
\usepackage[normalem]{ulem}
\usepackage{epstopdf}
\usepackage{booktabs}
\usepackage{cite}
\usepackage{rotating}

\newcommand{\cancel}[1]{}

\title{\LARGE \bf
Multi-lane Cruising Using Hierarchical Planning and Reinforcement Learning
}

\author{Kasra~Rezaee, Peyman~Yadmellat,  Masoud~S.~Nosrati, \\
Elmira~Amirloo~Abolfathi, Mohammed~Elmahgiubi, and Jun~Luo 
\thanks{Noah's Ark Lab., Huawei Technologies Canada, Markham, Ontario, Canada L3R 5A4}
}

\begin{document}

\maketitle
\thispagestyle{empty}
\pagestyle{empty}

\input{sections/abstract.tex}

\input{sections/introduction.tex}

\input{sections/related_works.tex}
\input{sections/technical_approach.tex}

\input{sections/simulator_experimental_design.tex}
\input{sections/experiments.tex}
\input{sections/conclusion.tex}

\bibliographystyle{IEEEtran}
\bibliography{bibliography} 

\end{document}

%% file: sections/abstract.tex
\begin{abstract}
Competent multi-lane cruising requires using lane changes and within-lane maneuvers to achieve good speed and maintain safety.
This paper proposes a design for autonomous multi-lane cruising by combining a hierarchical reinforcement learning framework with a novel state-action space abstraction.
While the proposed solution follows the classical hierarchy of behavior decision, motion planning and control, it introduces a key intermediate abstraction within the motion planner to discretize the state-action space according to high level behavioral decisions.
We argue that this design allows principled modular extension of motion planning, in contrast to using either monolithic behavior cloning or a large set of hand-written rules.
Moreover, we demonstrate that our state-action space abstraction allows transferring of the trained models \textit{without retraining} from a simulated environment with virtually no dynamics to one with significantly more realistic dynamics.
Together, these results suggest that our proposed hierarchical architecture is a promising way to allow reinforcement learning to be applied to complex multi-lane cruising in the real world.
\end{abstract}

%% file: sections/introduction.tex
\section{INTRODUCTION}
Developing autonomous cars that reliably assist humans in everyday transportation is a grand research and engineering challenge.
While autonomous cars are on the way to revolutionize the transportation system by increasing safety and improving efficiency~\cite{schoettle2014survey}, many aspects of driving remain beyond the reach of current solutions.
Mundane as it may seem, cruising on a multi-lane highway effectively and safely while taking full advantage of available driving space has proved challenging for existing autonomous cars.
What makes multi-lane cruising significantly more challenging than the single-lane adaptive cruise control (ACC) is the fact that the multi-vehicle interaction happens both laterally (i.e. perpendicular to the lanes) and longitudinally (i.e. parallel to the lanes) and requires coordination between lateral and speed control.
In particular, multi-lane cruising involves changing lanes, bypassing in-lane small objects, speed control, and maintaining safe distance from vehicles ahead.

The current research focuses on the use of hierarchical reinforcement learning for multi-lane cruising as a special case of driving on structured roads.
Driving on structured roads is heavily regulated by signs, signals, and rules that come to apply at various points in time and space.
In multi-lane cruising, lane markings dictate that driving takes place mostly within the boundaries of a single lane.
Lane change is a short-lived, transitional event in continuous motion space that links two distinct states~\textemdash~driving in one lane vs. driving in an adjacent lane.
Similarly, traffic rules\footnote{Traffic lights, lane markings, speed limit signs, fire truck sirens, etc.} are \textit{symbolically punctuated} states that can be viewed as a hierarchical planning system, through which higher level decisions on discrete state transitions are coordinated with lower level motion planning and control in continuous state space.
In this context, the hierarchical planning system is divided into three sub-systems: a) behavioral planner (BP), b) motion planner (MoP), and c) motion controller.
The BP is responsible for high level decision making (e.g. \textit{switch to the left lane}) over discrete states.
The MoP generates a continuous trajectory given behavioral commands.
The motion control module controls the vehicle to follow the planned trajectory.

Classical methods for implementing the BP are largely rule-based with finite state machines being a common choice~\cite{paden2016survey}.
Classical MoP methods typically require optimization according to explicitly defined cost functions with the behavior decision expressed as constraint terms in the cost function~\cite{werling2010optimal,hu2018dynamic}.
Rule-based BP is extremely hard to maintain and does not scale well in complex dynamic scenarios.
Likewise, explicit cost functions for MoP are hardly general enough and very difficult to tune for complex dynamic interactions.
These limitations could explain the conservative behavior of current autonomous cars in multi-lane driving.

In response to these limitations, many recent studies attempted learning-based approaches.
Bojarski~\textit{et al.}~\cite{bojarski2016end} proposed an end-to-end supervised learning scheme that directly maps images to steering commands.
Sun~\textit{et al.}~\cite{sun2017fast} in contrast use a mapping from state features to trajectories and then use an execution layer to further guarantee short term feasibility and safety.
These approaches leverage expert experience for training.
However, by directly cloning the expert's driving strategy, they are limited to the expert's performance and experience, failing to adequately explore the parts of the state-action space that may be less critical for safety and performance.
In addition, planning and control are largely implemented as one monolithic network, which makes debugging, failure analysis, and incorporation of domain knowledge all very difficult.

In contrast to end-to-end solutions, we tackle the problem through a hierarchical and modular scheme by breaking the multi-lane cruising problem into multiple distinct sub-tasks and providing separate modules to address each sub-task.
In our design, the MoP is separated into multiple motion planning submodules specialized for a driving sub-task (\textit{lane keeping, lane switching}).
The BP determines which motion planning submodule is required to be triggered at each time step.
Such design allows for
a) reducing the complexity in generating inclusive scenarios by focusing on task-specific scenarios;
b) achieving more efficient training by considering task-specific state-action representation and reward design;
and c) enabling ease of transfer through hierarchical and modular design.
Moreover, the motion controller in our design is realized through classical and none-learning based approaches to enable further transferability from simulation to real-world vehicles.

To summarize, the main contributions of this paper are: 
\begin{itemize}
\item proposing a modularized skill-based planning framework with two layers of hierarchy (behavioral and motion planner) for cruising in multi-lane roads;
\item  proposing a higher level of abstraction in the state-action space of driving in multi-lane roads.
\end{itemize}

In Section~II, we review the related methods in autonomous driving.
In Section~III, we present the details of our planning  framework.
Section~IV describes the simulation environment used for training and validation.
In Section~V we evaluate our approach comprehensively and conclude our work in Section~VI.

%% file: sections/related_works.tex
\section{RELATED WORKS}
Recent studies have utilized reinforcement learning (RL) for high-level decision making~\cite{mukadam2017tactical, mirchevskareinforcement}.
Mukadam~\textit{et al.}~\cite{mukadam2017tactical} proposed a Q-learning based approach to address the lane switching problem in autonomous driving.
A Q-network was considered to issue discrete high level commands, \textit{e.g.} switch left/right.
Mirchevska~\textit{et al.}~\cite{mirchevskareinforcement} proposed an RL-based approach for autonomous driving in highway scenarios using the fitted Q-iteration with extremely randomized trees as a function approximator.
Both of these approaches have utilized RL for high level decision making (\textit{i.e.} BP) and adopted classical and rule-based approaches for motion planning~\cite{mukadam2017tactical, mirchevskareinforcement} and collision avoidance~\cite{mukadam2017tactical}.

Wulfmeier~\textit{et al.}~\cite{wulfmeier2016watch} utilized inverse reinforcement learning to deduce human driver's underlying reward mapping  from sensory input by applying Maximum Entropy to a large-scale human driving dataset.
The deduced reward mapping was then used as a cost-function for motion planning.
The approach however focuses on static environment, and it is not directly applicable to environments involving dynamic obstacles (\textit{e.g.} multi-lane urban roads and highway).

A planning by prediction paradigm was proposed in~\cite{shalev2016long} to tackle adaptive cruise control and roundabout merging problems.
The navigation problem was decomposed into two prediction and planning phases.
In the prediction phase, supervised learning was utilized to predict the near future states based on the current states.
Then, RL was used in the planning phase to model the vehicle's acceleration given the prediction results.

A hierarchical reinforcement learning scheme was incorporated in~\cite{paxton2017combining} to deal with the self-driving problem in challenging environments.
The proposed scheme was formulated by decomposing the problem into a set of high level temporal-constrained options and low-level control policies, and using Monte Carlo Tree Search over the available options to find the best sequence of options to execute.
The main difficulty with realizing temporal-based methods stems from ambiguity on setting the termination condition.
The typical solution is to assign a fixed expiration time to each option and penalize the agent if execution time is expired.
Specifying such deadlines are a challenging and conflicting task that adds to the complexity of the overall training process.
For example, if the goal is to learn switch lane option, the operator requires to specify a deadline for completing the lane change.
If the selected deadline is too short, the agent may prioritize a sharp and unsafe lane change over an accident-free and smooth maneuver.
Similarly, extending the deadline may result in conservative or undesired behaviors.

Furthermore, most existing approaches rely on learning-based low-level control policies.
In practice, low-level policies may result in oscillatory or undesirable behaviors when deployed on real-world vehicles due to imperfect sensory inputs or unmodeled kinematic and dynamic effects.
Given well-established controllers such as PID and MPC, we believe that learning-based methods are more effective in the high and mid level decision making (\textit{e.g.} BP and MoP) rather than low-level controllers.

%% file: sections/technical_approach.tex
\section{TECHNICAL APPROACH}
\subsection{The Planning Hierarchy}
Driving is a symbolically punctuated behavior.
Different from regular robotic problems, driving is heavily punctuated by signs and rules on top of what is largely a continuous control task.
To name some, the symbols here include lane markings, traffic lights, speed limit signs, fire truck sirens, and the turning signals of other vehicles.
As an example, lane markings dictates that most driving happen within a single lane.
Thus, lane changes are short-lived and transitional events that link forward driving in one lane to forward driving in an adjacent lane -- two discrete states at a higher level of abstraction in the state space of driving.
Because driving is symbolically punctuated, it is naturally hierarchical:
higher level decisions on discrete state transitions with lower level execution in continuous state space, which suggests a hierarchical structure in the design of planning systems for autonomous driving.

\begin{figure}[]
    \centering
    \includegraphics[width=0.35\textwidth]{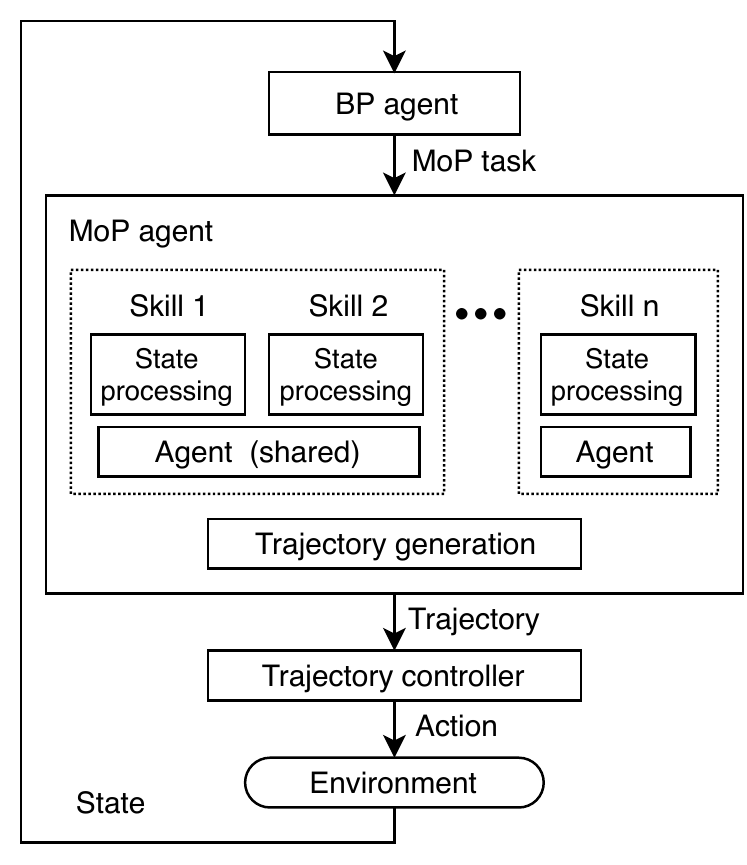}
    \caption{Overview of our hierarchical planning framework: Behavioral planner (BP) and Motion planners (MoP).}
    \label{fig_architecture}
\end{figure}

Figure~\ref{fig_architecture} illustrates our proposed hierarchical decision making architecture for cruise in multi-lane roads.
The proposed decision making framework includes \textit{BP} that makes high level decisions about transitions between discrete states, and \textit{MoP} that generates a target spatio-temporal trajectory with a target speed  according to the decisions made by BP.
The target trajectory is then fed to the controller  to follow the trajectory by controlling the steering wheel, throttle, and brake in continuous state space.

The hierarchical structure of our planning framework facilitates analysis of the decisions that are made during driving.
In addition, the structure allows for convenient modularization of different \textit{skills}, \textit{e.g.} adaptive cruise control, lane switching, pullover, and merging.
Each modularized skill acts as an independent entity and forms a comprehensive maneuver function considering its own constraints and safety internally.
This also enables modifying and replacing sub-modules according to new requirements and conditions.
Moreover, these modules can be shared among two or more driving sub-tasks to facilitate faster learning and generalization.

\subsection{Behavior Planner}
The behavior decision is about transitioning between states that are discrete only at a higher level of abstraction.
BP is responsible to drive the car to the  destination safely and as fast as possible.
In our current setting, BP makes high level decisions including \textit{keep lane}, \textit{switch to the left lane}, and \textit{switch to the right lane} subject to the following conditions:
\begin{itemize}
    \item navigating the ego-car to less busy lanes so the car can drive to the maximum speed limit (\textit{drive as fast as possible})
    \item avoiding collisions (\textit{drive safely}).
\end{itemize}

BP takes the full set of states as input which includes: ego lane, ego speed, distance and relative speed of the nearest vehicles in the front and back for current and neighboring lanes.
We design a \textit{coarse-grained} reward function and avoid any \textit{fine-grained} rules in our reward feedback.
This way, we give the RL agent a chance to explore the state space and to come up with solutions that possibly outperform classical rule-based planners.
The BP module receives a reward of $1$ if speed is above a threshold.
The threshold is higher for left lanes and lower for right lanes to motivate the agent to keep right.
A penalty of $-5$ is given for each left lane change.
Therefore, rewarding the agent for staying in a lane where it can drive fast, and discouraging excessive lane changes.
Otherwise, the BP agent's reward is $0$.
The BP reward can be summarized as:
\begin{equation}
r_{BP}=
\begin{cases}
-5, & lane(t) > lane(t-1) \\
1, &  speed(t) > threshold(lane(t)) \\
0, & \text{otherwise}
\end{cases}
\end{equation}
where $lane(t)$ being the lane number, starting from the rightmost lane, and increasing as we move towards left.
As mentioned above, the $threshold(lane(t))$ is higher for left lanes.

\subsection{Motion Planner}
Motion Planner's main task is to provide a safe and collision-free path towards its destination, while taking into account road boundaries, the vehicle kinodynamic constraints, or other constraints dictated by BP. 
In our design, the MoP generates a target trajectory defined as a set of 2D points (\textit{i.e.} path) coupled with a target \textit{speed} value.

\begin{figure}[]
    \centering
    \includegraphics[width=0.45\textwidth] {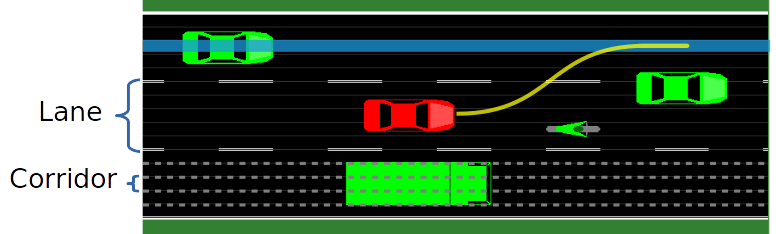}
    \caption{Corridor abstraction for structured roads.
    Here the \textit{Blue line} shows the corridor selected by MoP, and the \textit{Yellow line} is a generated trajectory corresponding to the selected corridor. The ego-car is shown with red color.}
    \label{fig_simulator}
\end{figure}

We propose a new level of road abstraction, through which each lane consists of $N_c$ corridors, as depicted in Figure~\ref{fig_simulator}.
Corridors are defined in the Fren{\'e}t coordinate frame parallel to the desired trajectory.
This trajectory is constructed based on road centers (waypoints) or path planning algorithms for unstructured environments.
As corridors are defined in the Fren{\'e}t coordinate frame, tracking performance remains invariant to transformation \cite{werling2010optimal}.

An MoP agent in our framework selects two sets of actions: 1) A \textit{lateral action} identifying the target corridor; and 2) A \textit{speed action} which selects the target speed.
Corridor selection is equivalent to selecting a path among a set of predefined paths (clothoids or splines).

The expected behavior of the MoP module differs with respect to the BP action:
\begin{itemize}
    \item For \textit{keep-lane}, MoP should select a corridor within the current lane while avoiding collision.
    This enables MoP to maneuver around small objects in the lane without switching lane.
    \item For \textit{switch-left/right}, MoP should select a corridor towards the direction of lane change. The MoP agent is also allowed to select corridors within the current lane to avoid collision when the target lane is occupied.
\end{itemize}

Speed set-point is selected according to BP actions and physical boundaries (\textit{e.g.} heading cars, or any interfering objects on the road) while ensuring safety and smoothness.
Simultaneously adjusting the speed set-point and lateral position allows to handle various tasks such as in-lane cruising, emergency stop, and vehicle following.
This also includes any other speed-dependent BP actions.

\begin{figure}[]
    \centering
    \includegraphics[width=0.4\textwidth]{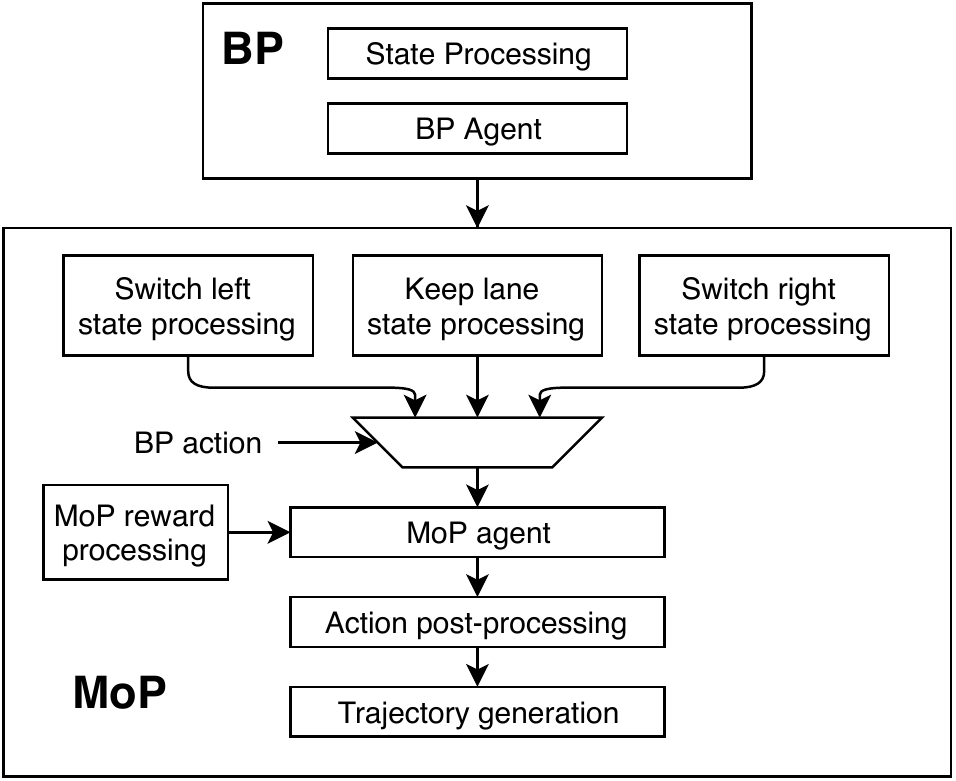}
    \caption{Diagram showing the highway cruising problem implemented in the hierarchical RL framework.}
    \label{fig_arch_implementation}
\end{figure}

Figure~\ref{fig_arch_implementation} shows an overview of the hierarchical framework applied to the highway cruising problem. The keep-lane and switch-lane tasks are achieved using a shared MoP agent. Given BP action, the corresponding pre-processing module is triggered. Each pre-processor passes a relevant subset of states to the shared agent. The selected target corridor and speed set-point by the MoP agent are relative quantities. The absolute corridor and speed values are calculated in the \textit{Action post-processing} module, and fed into the \textit{Trajectory generator} module. Trajectory generator is a non-learning-based module implemented using a simple curve fitting from point $A$ to point $B$ (Yellow line in Figure~\ref{fig_simulator}). The generated trajectory is extended along the target corridor as the vehicle moves and is guaranteed to be smooth and feasible for the controller node.

The input states to the shared MoP agent include current speed, speed set-point, BP target speed, and the current corridor.
In addition, the front and back gaps along with their rate of change are also used for a number of corridors equivalent to one lane.
The shared agent outputs a new corridor relative to current corridor and change in speed set-point.

The MoP agent receives a reward of $1$ if it is moving close to the BP's target speed or following another vehicle with a safe distance $d$. 
The distance $d$ is defined as $d=v\times\tau+d_0$, where $v$ is the ego speed, $\tau$ is desired headway in seconds, and $d_0$ is the distance when stopped.
The reward for safe following is only awarded if all the corridors are blocked, i.e. the ego vehicle cannot bypass the obstacle with an in-lane maneuver.
Additionally, to promote driving in the center of the lane, the agent is only awarded if the ego vehicle is in the middle corridor.

In summary, the following conditions need to be met for the MoP agent to receive a reward of $1$:
\begin{itemize}
    \item being in the middle corridor, AND
    \item EITHER the speed of the ego vehicle is within a threshold of the BP target speed,
    \item OR the minimum front gap is within a threshold of the safe distance $d$.
\end{itemize}
Otherwise, the reward that MoP agent receives is $0$. While not necessary, it is helpful to add a penalty term for high acceleration to direct the MoP agent to opt for less aggressive actions when possible; thereby, resulting in more comfortable driving experience.

For the keep-lane task, the MoP states are limited to corridors associated with the current lane.
The corridors on the sides that result in ego vehicle intruding other lanes are set to be blocked, by setting their gaps to $0$. This results in a collision effect if the ego vehicle enters those corridors.
For switch-lane tasks, the corridors are chosen such that the ego vehicle is off from the middle corridor by one corridor, with the corridor offset being opposite the direction of lane change.
Since the MoP agent only receives a reward in the middle corridor, this choice for corridors will direct the MoP agent to move toward the desired lane.
It is worth noting that the MoP agent is not forced to make the corridor change, \textit{e.g.} it can choose to remain off the middle corridor when the target corridor is occupied.
During a switch-lane task, as soon as the corridor of the ego vehicle changes, the state processing shifts the corridors again.
It is expected from BP to change the action to keep-lane when the ego vehicle arrives at the target lane.
If the BP action changes from switch-lane to keep-lane before the ego vehicle arrives at the new lane (canceling a lane change) the MoP also cancels the lane change process and return to the middle of the current lane.

\subsection{Training}
In the proposed hierarchical RL framework, BP issues a high level command which is executed by the corresponding MoP. As opposed to the other hierarchical frameworks (\textit{e.g.} \cite{kulkarni2016hierarchical}), BP does not wait until its command gets executed.
Considering any fixed lifetime for BP commands is dangerous for autonomous driving.
In fact, BP should be able to update its  earlier decisions (at every time step) according to the new states.
MoP is designed to prioritize safety over BP decisions.

Our framework is flexible when it comes to choosing RL algorithms to be applied for BP and MoPs. We tested our framework with \textit{DQN} \cite{mnih2015human}.
The training was carried out by training MoP agents to achieve certain performance and reliability in executing the sub-tasks, with a random BP module changing the target lane every few hundred steps.
Then, BP agent was trained using the trained MoP agent.
This allows for the BP agent to learn the weakness and strength of the MoP agents, and potentially improving the overall performance of the planning system.

%% file: sections/simulator_experimental_design.tex
\section{SIMULATOR}
For training of the BP and MoP RL agents, we employed the SUMO traffic simulation software \cite{SUMO2018}.
A wrapper, called \textit{gym-SUMO} was developed in Python to make an OpenAI Gym compatible environment for training.
While the ego vehicle is not limited to the center of the lane and can freely move using gym-SUMO, the default behavior for other vehicles in SUMO is confined to the center of lanes.
To make the traffic more realistic with richer corridor information, we trained the agents with the \textit{sublane} model activated in SUMO.
Sublane model in SUMO provides more natural and gradual lane change behavior.

Given that SUMO has built-in speed and lateral controls and does not have sophisticated kinematic and dynamic models, the action inputs were defined as the desired speed and the target corridor.

The output from gym-SUMO includes state variables required by the MoP and BP agents, the BP agent reward, and a termination signal for accidents.
The reward of MoP agent is calculated internally from the environment state.

To evaluate the full architecture with a more realistic environment we employed Webots \cite{Webots} in conjunction with SUMO.
The ego vehicle is controlled through Webots providing a realistic dynamic and kinematic model, while SUMO controls the rest of traffic movement providing an intelligent and realistic environment.

\begin{figure}[!htb]
      \centering
      \includegraphics[height=8cm, clip]{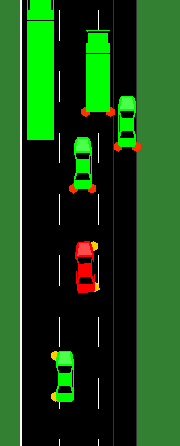}
      \includegraphics[height=8cm, clip]{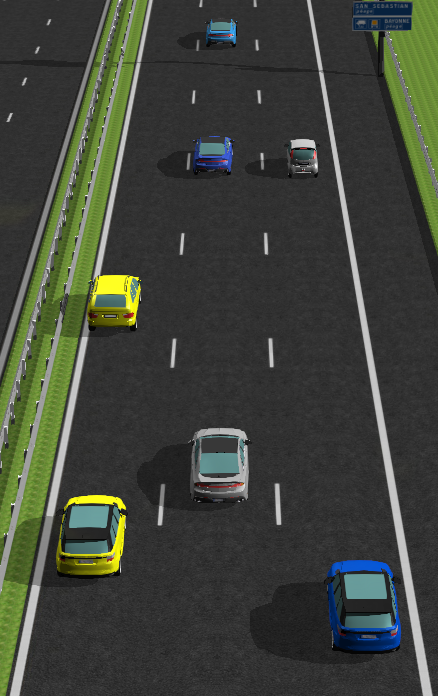}
      \caption{screenshots of gym-SUMO (left) and Webots (right) simulation environments. Screenshots are not from the same timestep.}
      \label{fig_sumo_webots}
\end{figure}

The road network used for training and evaluation consisted of three lanes in a straight line.
Figure~\ref{fig_sumo_webots} depicts a snapshot of the gym-SUMO and Webots environments employed for evaluation.

%% file: sections/experiments.tex
\begin{figure*}
    \centering
    \includegraphics[width=0.325\textwidth, trim={0cm 0cm 0cm 0cm}]{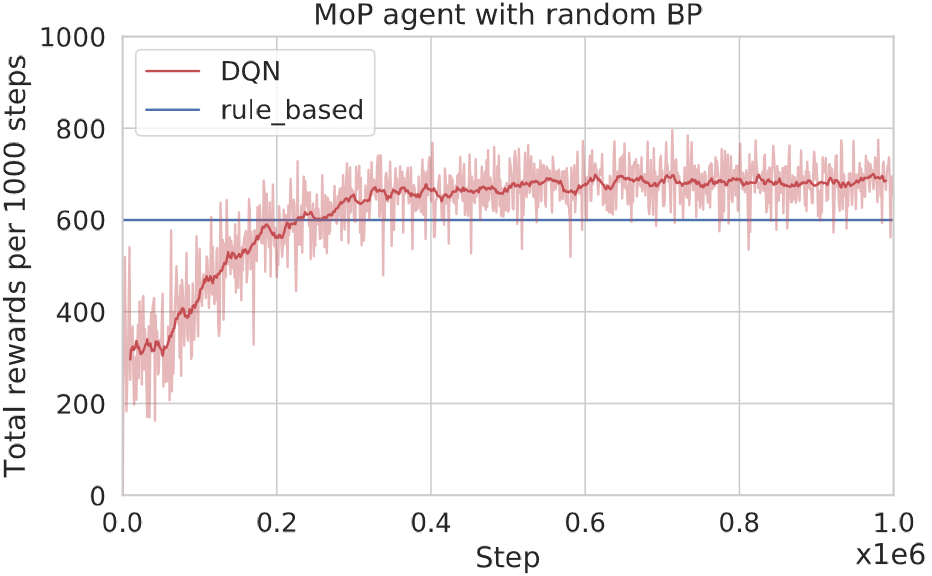}
    \includegraphics[width=0.325\textwidth, trim={0cm 0cm 0cm 0cm}]{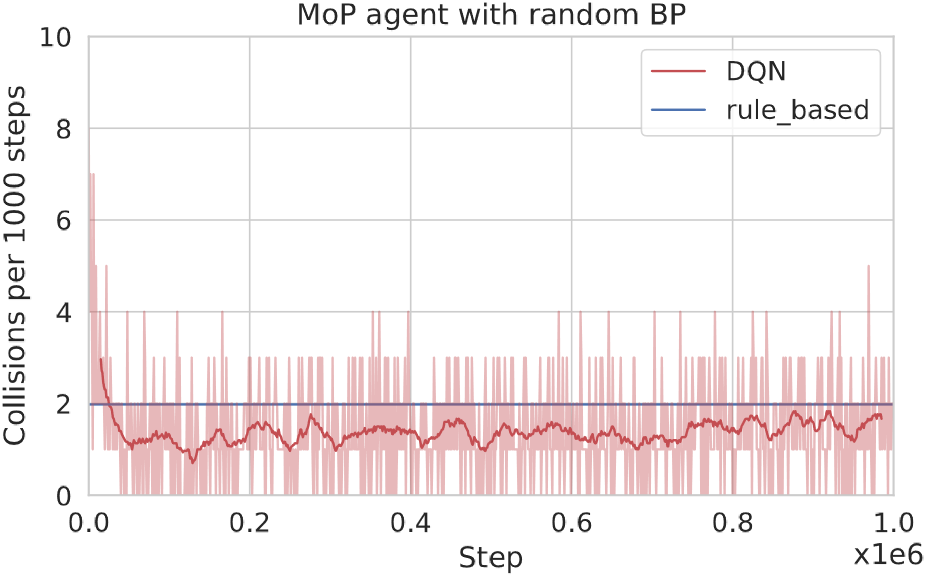}
    \includegraphics[width=0.325\textwidth, trim={0cm 0cm 0cm 0cm}]{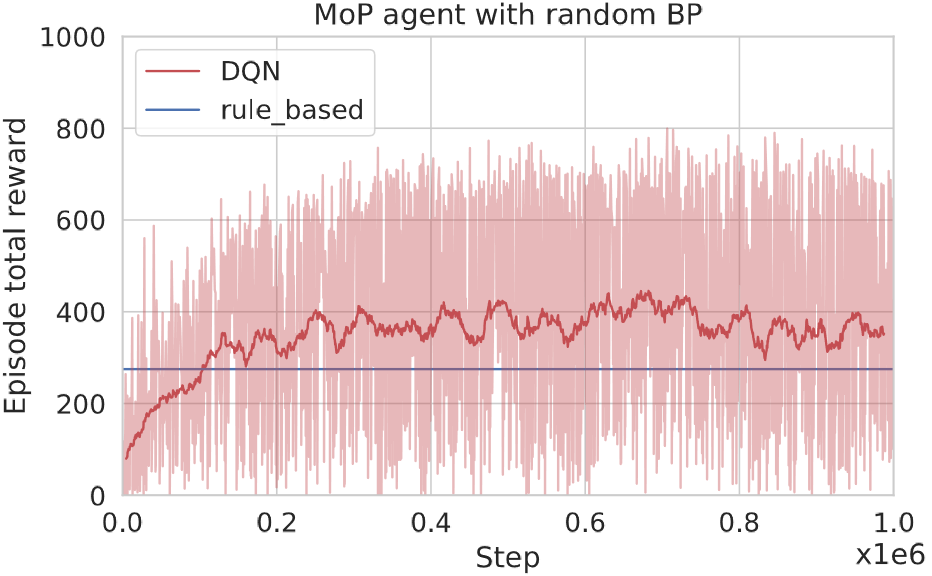}
    \caption{Training evolution of the MoP agent (darker line show the trend with a moving average) (left) total reward per 1000 steps, (center) average collisions per 1000 steps, (right) total reward per episode (1000 steps or collision).}
    \label{fig_mop_train}
\end{figure*}

\section{EXPERIMENTS}
In this section we present a set of experiments to evaluate the performance of our method to learn a policy for automated highway cruising.
We chose a popular Q-learning-based technique, namely DQN, to test our framework.
Additionally, a set of rule-based BP and MoP were developed to provide a baseline for comparison.
The rule-based algorithms were developed so that they achieve similar goals as the reward defined for the RL-based BP and MoP.
While the rule-based algorithms are by no means optimal, every effort was made to have the best performing algorithms.

\subsection{Gym-SUMO}
We trained the MoP and BP agents for $1$ million and $400$ thousand steps, respectively, using the gym-SUMO simulator.
We first trained the MoP agent with a BP agent that requests a lane change every $100$ steps.
Figure~\ref{fig_mop_train} show the training evolution of the MoP agent.
The darker line shows the trend of values for the learning agent with moving average.
The horizontal line shows the performance of the rule-based approach averaged over $40000$ steps.
The agent is clearly learning as the average reward is increasing and number of collisions is decreasing.
However, they do not show the whole picture on their own as high average reward can be due to high speed and higher tendency to collide with other vehicles.
The overall performance of the agent can be captured by the episode total reward as shown in Figure~\ref{fig_mop_train}(right).
An episode terminates either with a collision or after $1000$ steps.

\begin{table*}[h]
\caption{Comparison results between hierarchical RL and rule-based agents for BP task.}
\label{table_bp_perf}
\begin{center}
\begin{tabular}{@{}lcccc@{}}
\toprule
 & Avg. BP reward & Avg. num. of collisions &  Avg. num. of lane changes & Avg. speed [kph] \\ \midrule
Hierarchical RL            & 699 & 2.125 & 25.0 & 43.21 \\
Rule-based BP and MoP      & 627 & 3.525 & 73.0 & 44.24 \\
Rule-based BP with DQN MoP & 531 & 3.350 & 79.2 & 41.29 \\
SUMO controlling the ego   & 511 & 1.625 & 88.3 & 41.78 \\ \bottomrule
\end{tabular}
\end{center}
\end{table*}

\begin{figure}
    \centering
    \includegraphics[width=1.0\columnwidth,trim={0cm 0.5cm 0cm 0.4cm}, clip]{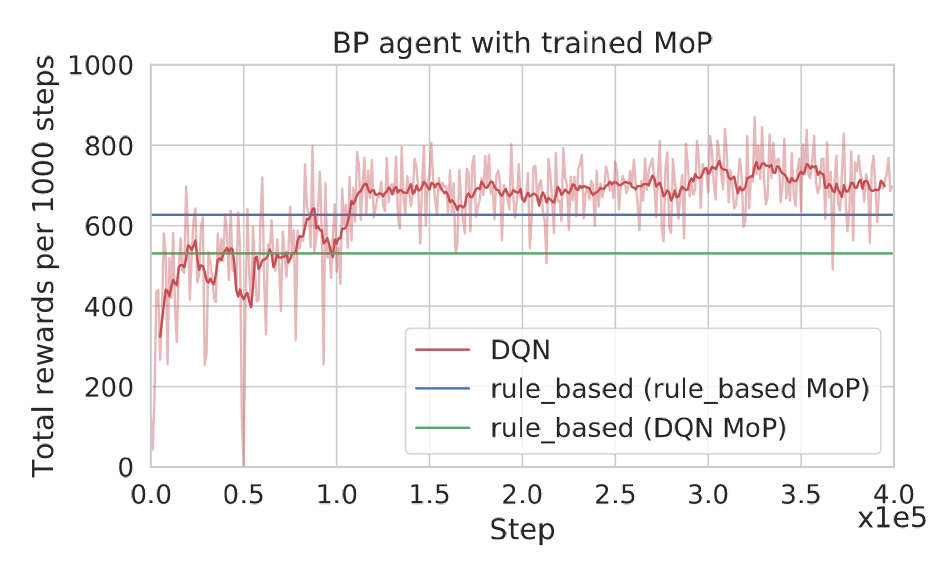}
    \caption{Training evolution of the BP agent.}
    \label{fig_bp_train}
\end{figure}

The training evolution of the BP agent is shown in Figure~\ref{fig_bp_train}.
Since the BP agent is trained with a previously trained MoP agent, its performance is relatively high from the start.
For the rule-based BP, we included both rule-based MoP and DQN-based MoP.
While the rule-based BP does not achieve good performance with DQN-based MoP, the DQN-based BP can adapt to MoP behavior and achieves much higher performance.

Note that the gym-SUMO simulation platform employed is fairly stochastic.
Therefore, there is significant variation in the rewards over time.
The variation present in the training evolution figures are due to this stochasticity rather than changes in behavior of the BP or MoP agents.

Table~\ref{table_bp_perf} summerizes the evaluation results for the trained agents together with those of baseline agents for 40000 steps. The result reported in the table are sums over 1000 steps to be comparable with the figures. We also incorporated the built-in SUMO model for further comparison.
As can be seen, the RL agent achieves higher average reward with lower average collisions compared to rule-based algorithms.
While the rule-based algorithms are designed to avoid collisions, collisions can also happen due to behavior of other SUMO vehicles.
This is evident from the average number of collision even when the ego is controlled by SUMO.
Observing the average speed, we can see that the RL agent achieves a comparable result with much lower lane changes, \textit{equivalent to higher comfort}.

\subsection{Webots}
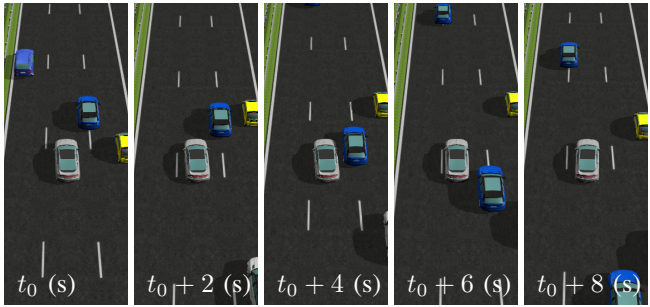
\begin{figure}[]
    \centering
    \setlength{\unitlength}{0.1\columnwidth}
    \begin{picture}(9.75,4.25)
        \put(0,0){\includegraphics[width=0.19\columnwidth, trim={9cm 9cm 16cm 6cm}, clip]{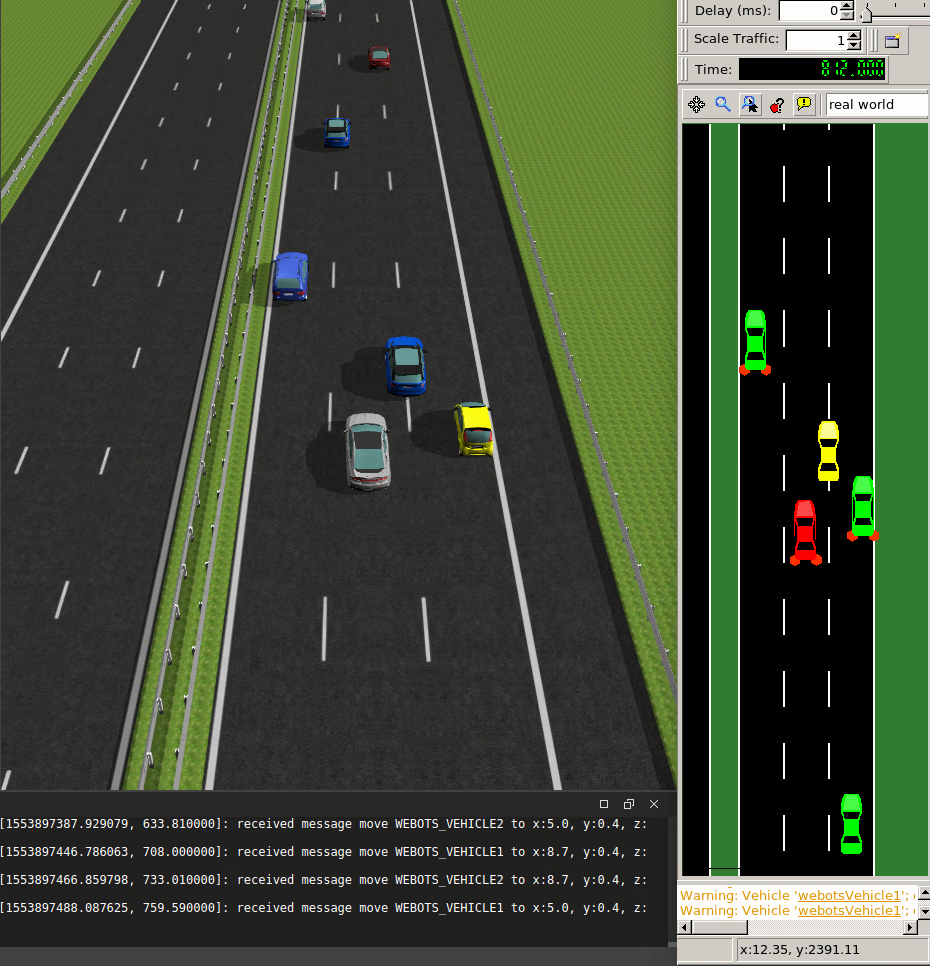}}
        \put(2,0){\includegraphics[width=0.19\columnwidth, trim={9cm 9cm 16cm 6cm}, clip]{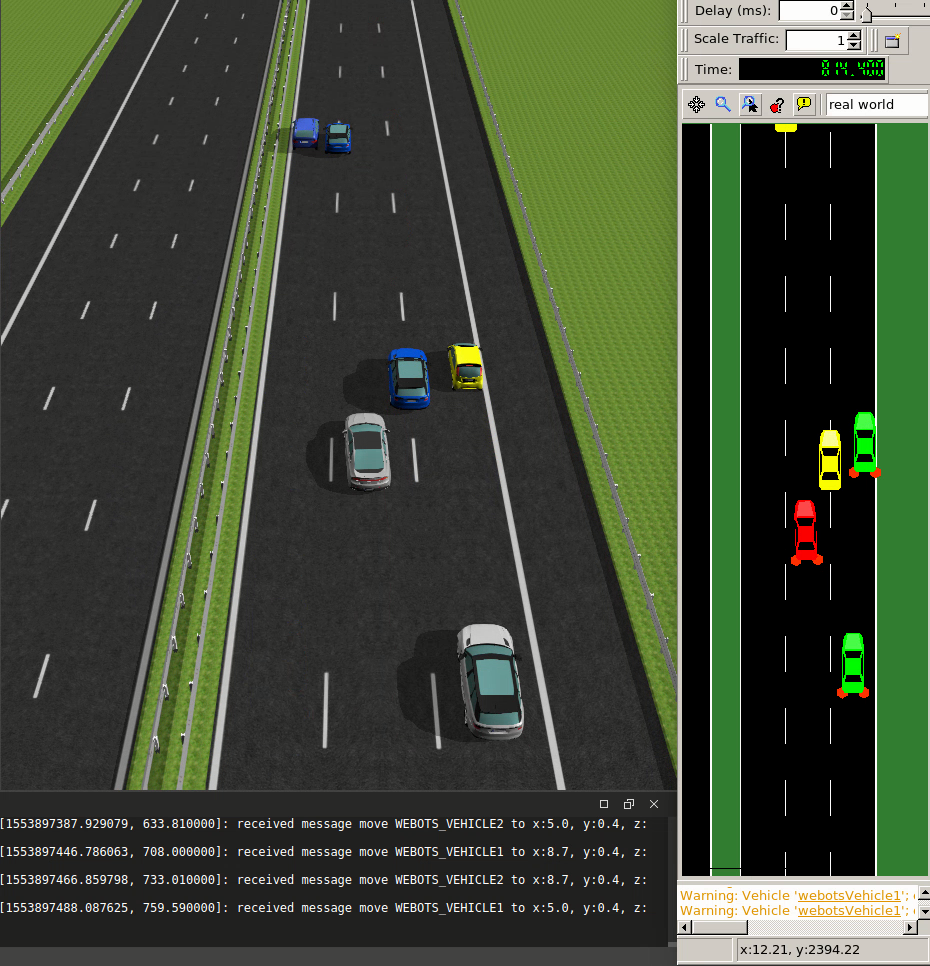}}
        \put(4,0){\includegraphics[width=0.19\columnwidth, trim={9cm 9cm 16cm 6cm}, clip]{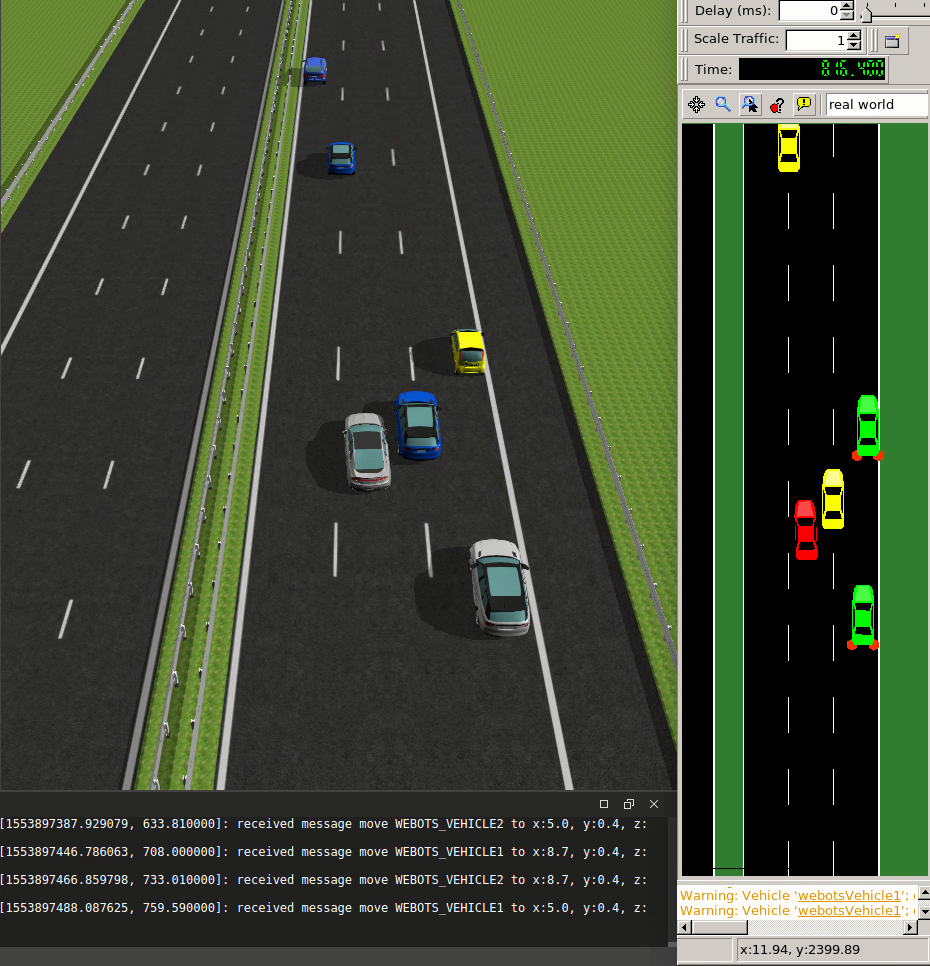}}
        \put(6,0){\includegraphics[width=0.19\columnwidth, trim={9cm 9cm 16cm 6cm}, clip]{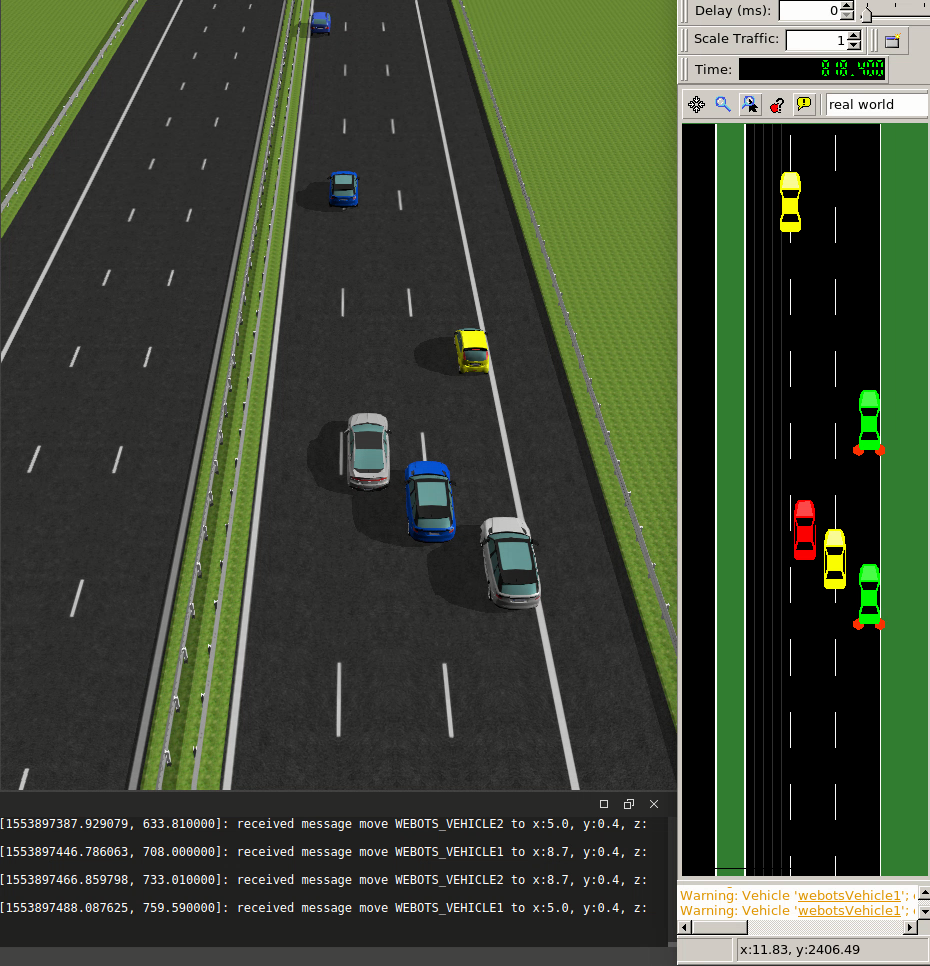}}
        \put(8,0){\includegraphics[width=0.19\columnwidth, trim={9cm 9cm 16cm 6cm}, clip]{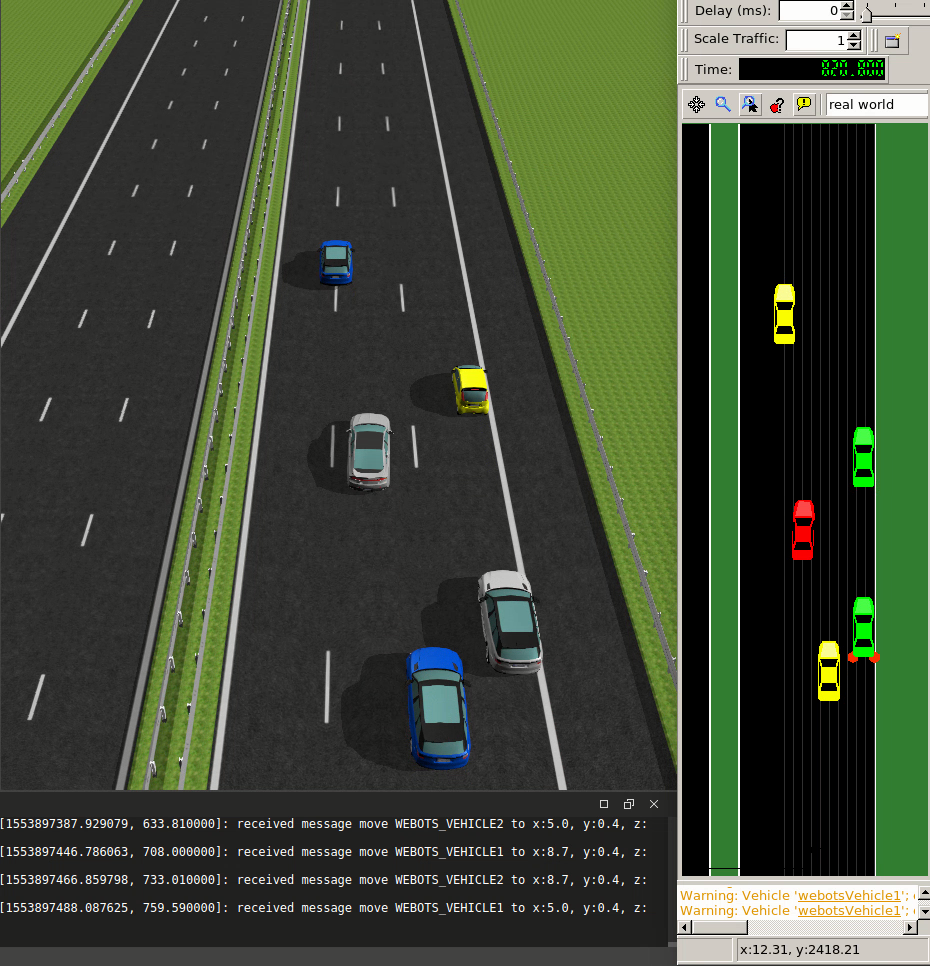}}
        \color{white}
        \put(0.2,0.2){$t_0$ (s)}
        \put(2.2,0.2){$t_0 + 2$ (s)}
        \put(4.2,0.2){$t_0 + 4$ (s)}
        \put(6.2,0.2){$t_0 + 6$ (s)}
        \put(8.2,0.2){$t_0 + 8$ (s)}
    \end{picture}
    \caption{Screenshots of Webots environment showing the ego vehicle bypassing a vehicle that is partially blocked the lane}
    \label{fig_webots_bypass}
\end{figure}

Webots provides a more realistic evaluation environment with relatively accurate vehicle dynamics and kinematics.
The traffic network used in Webots is identical to the one employed in gym-SUMO.
In Webots evaluations, we employed a timestep of $0.2$ sec to have smoother control.
Figure~\ref{fig_webots_bypass} shows screenshots of the ego motion while bypassing a vehicle partially blocking the lane in the webots environment.

\begin{figure}
    \centering
    \includegraphics[width=1.0\columnwidth, clip]{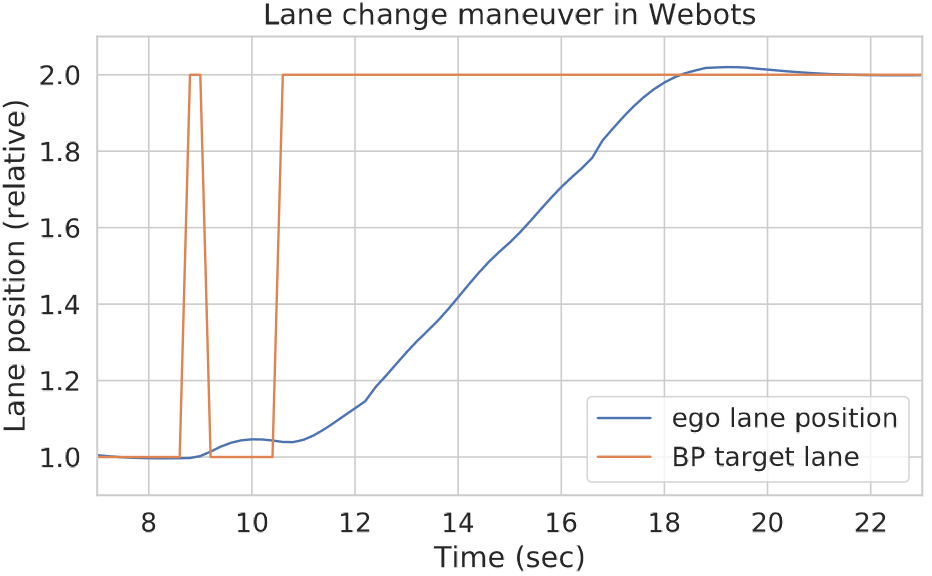}
    \caption{Ego vehicle making a lane change maneuver in Webots.
    The vehicle speed is $13.9~m/s$ during the lane change.}
    \label{webots_lane_change}
\end{figure}

\begin{figure}
    \centering
\setlength{\unitlength}{0.1\columnwidth}
\begin{picture}(10,5)
\put(0.5,0){\includegraphics[width=0.9\columnwidth, trim={0.85cm 0cm 0.8cm 1.1cm}, clip]{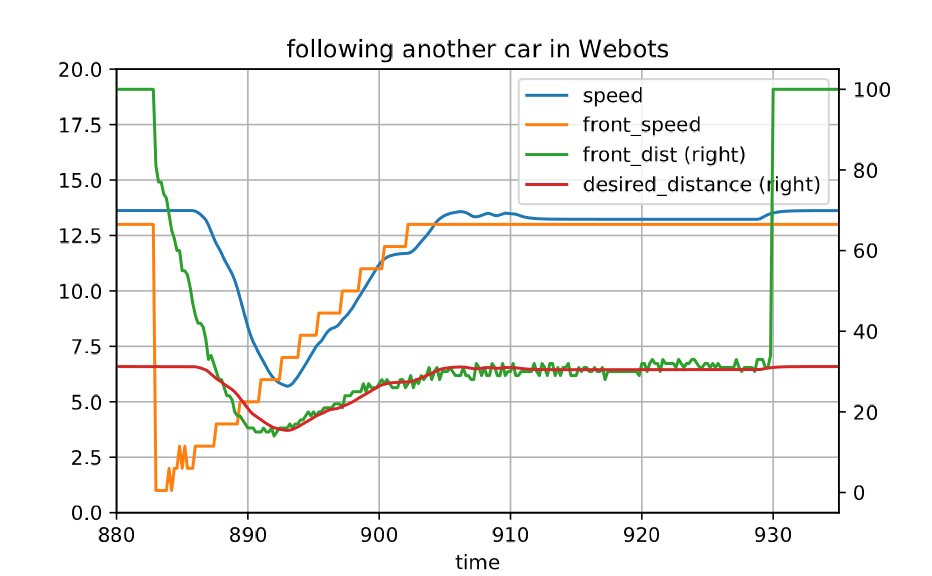}}
\put(0,2){\begin{turn}{90}Speed~($m/s$)\end{turn}}
\put(9.75,4.5){\begin{turn}{-90}Front gap~($m$)\end{turn}}
\end{picture}
\caption{Ego vehicle behavior while following a slower vehicle in Webots.}
\label{webots_follow}
\end{figure}

Figure~\ref{webots_lane_change} shows the lane change maneuver of the trained agent in Webots.
Note how MoP attempts to move back to original lane after BP temporarily cancels the lane change command.
The vehicle speed is $13.9~m/s$ ($50~kph$) during the lane change and in the absence of temporary cancellation lane change would have been completed in around $8$ sec.

Figure~\ref{webots_follow} shows the behavior of the trained agent while following a slower vehicle.
As can be seen at around $885~s$, the MoP anticipates the front vehicle getting too close and slows down to maintain safe distance.
The agent also smoothly speeds up as the vehicle in the front increase its speed.

%% file: sections/conclusion.tex
\section{CONCLUSIONS \& FUTURE WORK}
We proposed an RL-based  hierarchical framework for autonomous multi-lane cruising.
We introduced a key intermediate abstraction within the MoP to discretize the state-action space according to high level behavioral decisions.
Furthermore, we showed that the hierarchical design for an autonomous vehicle system can effectively learn the behavior and motion planner tasks.
The proposed framework allows for principled modular extension of motion planning, which is not the case in rule-based or monolithic behavior cloning-based approaches.
Moreover, we experimentally showed that our state-action space abstraction allows transferring of the trained models from a simulated environment with virtually no dynamics to the one with significantly more realistic dynamics without a need for retraining.

Although training BP and MoP individually could sufficiently address the cruising in multi-lane problem, as our future work, we aim to train the BP and MoP agents jointly (in an end-to-end fashion) to acquire higher level of performance.
Training BP and MoP in an end-to-end fashion helps both higher and lower levels to adapt to each other and potentially improves the overall performance.
Future works also include extending behavioral and motion planning capabilities (training more sub-tasks) to handle more driving situations, such as all-way stop signs and uncontrolled left turns.